\documentclass[final]{cvpr}

\usepackage{times}
\usepackage{epsfig}
\usepackage{graphicx}
\usepackage{subfigure}
\usepackage{amsmath}
\usepackage{amssymb}
\usepackage{booktabs}
\usepackage{bm}
\usepackage{algorithm}
\usepackage{algorithmic}
\usepackage{multirow}

\usepackage[pagebackref=true,breaklinks=true,colorlinks,bookmarks=false]{hyperref}

\def\cvprPaperID{4291} % *** Enter the CVPR Paper ID here
\def\confYear{CVPR 2021}

\begin{document}

%%%%%%%%% TITLE
\title{Effective, Efficient and Robust Neural Architecture Search}

\author{Zhixiong Yue\thanks{The first three authors contributed equally.}, Baijiong Lin, Xiaonan Huang, Yu Zhang\\
Department of Computer Science and Engineering, Southern University of Science and Technology\\
{\tt\small \{yuezx,linbj,huangxn3\}@mail.sustech.edu.cn,yu.zhang.ust@gmail.com}
% For a paper whose authors are all at the same institution,
% omit the following lines up until the closing ``}''.
% Additional authors and addresses can be added with ``\and'',
% just like the second author.
% To save space, use either the email address or home page, not both
%\and
%Second Author\\
%Institution2\\
%First line of institution2 address\\
%{\tt\small secondauthor@i2.org}
}

\maketitle

%%%%%%%%% ABSTRACT
\begin{abstract}
Recent advances in adversarial attacks show the vulnerability of deep neural networks searched by Neural Architecture Search (NAS). Although NAS methods can find network architectures with the state-of-the-art performance, the adversarial robustness and resource constraint are often ignored in NAS. To solve this problem, we propose an Effective, Efficient, and Robust Neural Architecture Search (E2RNAS) method to search a neural network architecture by taking the performance, robustness, and resource constraint into consideration. The objective function of the proposed E2RNAS method is formulated as a bi-level multi-objective optimization problem with the upper-level problem as a multi-objective optimization problem, which is different from existing NAS methods. To solve the proposed objective function, we integrate the multiple-gradient descent algorithm, a widely studied gradient-based multi-objective optimization algorithm, with the bi-level optimization. Experiments on benchmark datasets show that the proposed E2RNAS method can find adversarially robust architectures with optimized model size and comparable classification accuracy.
\end{abstract}

%%%%%%%%% BODY TEXT
\section{Introduction}

\begin{figure}[!htbp]
\centering
\includegraphics[width=\linewidth]{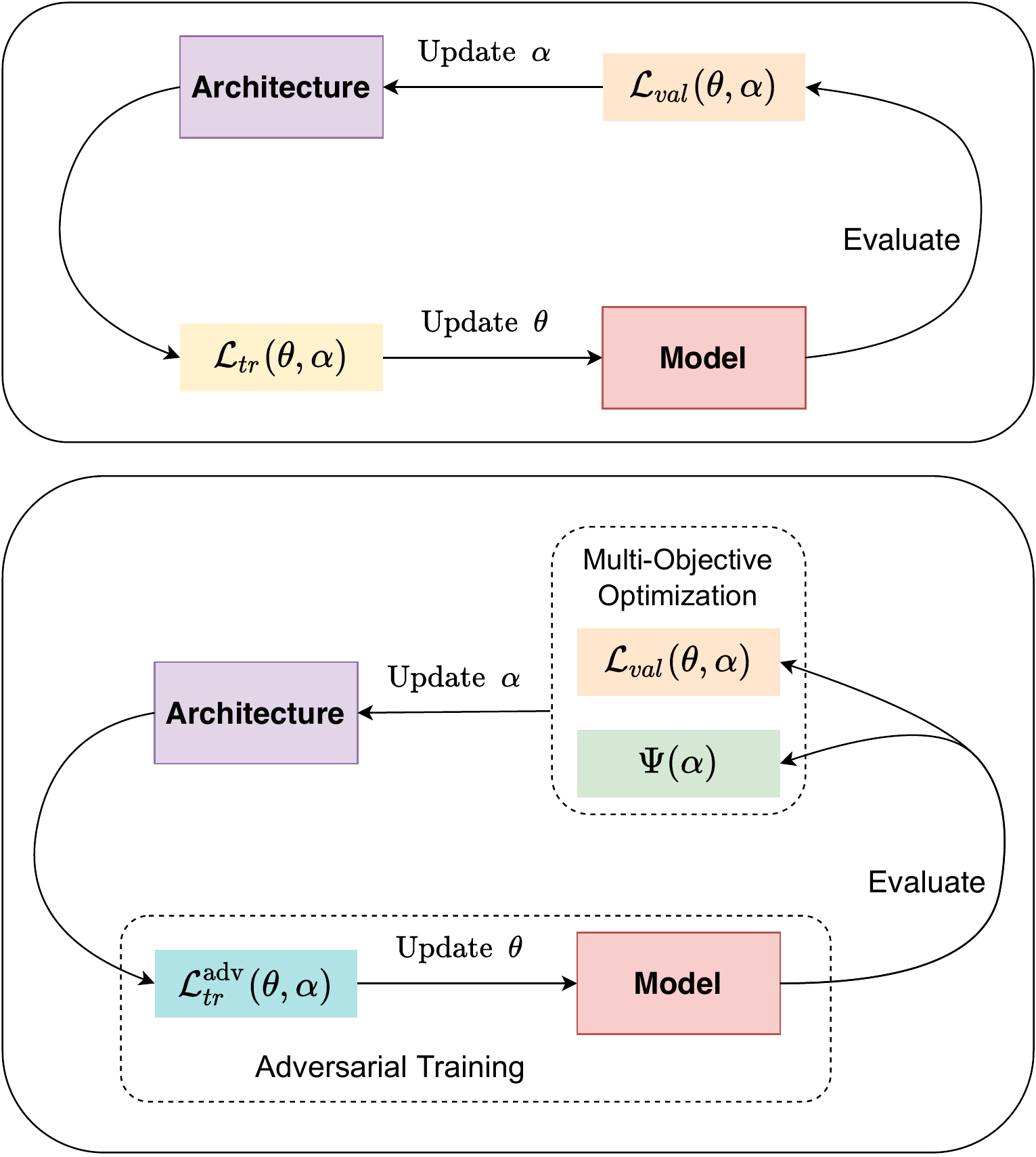}
\caption{Comparison of the architecture searching procedure between DARTS (top) and the proposed E2RNAS (bottom). We formulate E2RNAS as a bi-level multi-objective optimization problem. There are two key differences between E2RNAS and DARTS. Firstly, an adversarial training method is adopted to improve the robustness in the proposed E2RNAS model. Secondly, we evaluate the E2RNAS model with two objectives, including the validation loss $\mathcal{L}_{val}(\bm{\theta}, \bm{\alpha})$ and the number of parameters $\Psi(\bm{\alpha})$, for both the effectiveness and efficiency which is to learn a compact architecture with a controllable number of parameters. Therefore, E2RNAS can search an effective, efficient, and robust architecture.}
\label{fig:motivation}
\end{figure}

Deep learning has achieved great success in many areas, such as computer vision, nature language processing, speech, gaming and so on. The design of the neural network architecture is important for such success. However, such design relies heavily on the knowledge and experience of experts and even experienced experts cannot design the optimal architecture. Therefore, Neural Architecture Search (NAS), which aims to design the architecture of neural networks in an automated way, has attracted great attentions in recent years. NAS has demonstrated the capability to find neural network architectures with state-of-the-art performance in various tasks \cite{emh19, lsy19, tcpvshl19, xzll19}. Search strategies in NAS are based on several techniques, including reinforcement learning \cite{pham2018efficient, zoph2016neural}, evolutionary algorithms \cite{liu2017hierarchical, real2019regularized}, Bayesian optimization, and gradient descent \cite{lsy19, xu2019pc, chen2019progressive}. As a representative of gradient-descent-based NAS methods, the Differentiable ARchiTecture Search (DARTS) method \cite{lsy19} becomes popular because of its good performance and low search cost.

However, those NAS methods are typically only designed for optimizing the accuracy during the architecture searching process while neglecting other significant objectives, which results in very limited application scenarios. For example, a deep neural network with high computational and storage demands is difficult to deploy to embedded devices (\eg mobile phone and IoT device), where the resource is limited. Besides, the robustness of deep neural networks is also important. It is well known that the trained neural networks are easily misled by adversarial examples \cite{fgsm15, pgd17, fast_fgsm}, which makes them risky to deploy in real-world applications. For example, a spammer can easily bypass the anti-spam email filter system by adding some special characters as perturbations, and a self-driving car cannot recognize the guideboard correctly after sticking some adversarial patches.

Therefore, multi-objective NAS has drawn great attention recently because we need to consider more than performance when NAS meets real-world applications \cite{cztl20, emh19, tcpvshl19}. In \cite{jin2019rc, emh19, cztl20}, the model size and computational cost are considered to satisfy some resource constraint. Besides, some works \cite{Dong20, RobNets} search for differentiable architectures that can defense adversarial attacks. However, to the best of our knowledge, there is no work to simultaneously optimize the three objectives, \ie the performance, the robustness, and the resource constraint.
% To summarize, it remains an open question of how to search differentiable architectures that satisfy multiple objectives.

To fill this gap, this paper proposes an \textbf{E}ffective, \textbf{E}fficient, and \textbf{R}obust \textbf{N}eural \textbf{A}rchitecture \textbf{S}earch method (E2RNAS) to balance the trade-off among multiple objectives. Built on DARTS, the proposed E2RNAS method formulates the entire objective function as a bi-level multi-objective optimization problem where the upper-level problem is a multi-objective optimization problem, which can be viewed as an extension of the objective function proposed in DRATS. To the best of our knowledge, there is little work to solve such bi-level multi-objective optimization problem based on gradient descent techniques. To solve such problem, we propose an optimization algorithm by combining the multiple gradient descent algorithm (MGDA) \cite{desideri12} and the bi-level optimization algorithm \cite{colson2007overview}.
% \footnote{***Add a reference here.}. %Firstly, we integrate the robustness objective into the inner loop to optimize with the training objective function. Then, we use MGDA to find an efficient solution for minimizing model size while keeping comparable accuracy.

Specifically, the contributions of this paper are three-fold.
\begin{itemize}
\vspace{-0.3cm}
\item
We propose the E2RNAS method for searching effective, efficient and robust network architectures, leading to a practical DARTS-based framework for multi-objective NAS.
\vspace{-0.3cm}
\item
We formulate the objective function of the E2RNAS method as a novel bi-level multi-objective optimization problem and propose an efficient algorithm to solve it.
\vspace{-0.3cm}
\item Experiments on benchmark datasets show that the proposed E2RNAS method can find adversarially robust architectures with optimized model size and comparable classification accuracy.
\end{itemize}

\section{Related Works}

% \cite{emh19,tcpvshl19,cztl20}

\subsection{Adversarial Attack and Defence}
Deep neural networks are not robust while facing adversarial attacks \cite{sze14}. Most adversarial attacks are white-box attacks that assume attack algorithms can access to all configurations of the trained neural network, including the architecture and model weights. %Many of these white-box attack algorithms generate adversarial examples by calculating the model gradients.

% Given a neural network parameterized
% \begin{equation}
%     \sum_{i}\max_{\|\delta\|\leq\epsilon} \ell(\theta, x_i+\delta, y_i)
% \end{equation}

\vspace{-0.35cm}
\paragraph{Fast Gradient Sign Method}
Goodfellow \etal~\cite{fgsm15} propose Fast Gradient Sign Method (FGSM) for generating adversarial examples. It directly uses the sign of the gradient of the loss function with respect to weights as the direction of the adversarial perturbation as
% Then, they change all pixels with a small magnitude on the direction to generate the adversarial example as
{\small
\begin{equation*}
\bm{\delta}=\epsilon\cdot\mathrm{sign}(\nabla_\mathbf{x} \ell\left(\bm{\theta}, \mathbf{x}, y\right)),
\end{equation*}}
% where $x^\prime$ is the generated adversarial example,
% where $\delta$ is the initial value of the adversarial perturbation that satisfies $\left\|\delta\right\|_{\infty}\leq\epsilon$, $\|\cdot\|_{\infty}$ denotes the $\ell_{\infty}$ norm,
where $\mathbf{x}$ is the original input, $\epsilon$ is a small scalar to represent the strength of the perturbation, $\bm{\theta}$ denotes parameters of the victim model, $y$ is the original ground-truth label for input
$\mathbf{x}$, $\mathrm{sign}(\cdot)$ denotes the elementwise sign function, $\ell(\cdot, \cdot, \cdot)$ denotes the loss function used for training the victim model, and $\nabla_{\mathbf{x}}\ell\left(\bm{\theta}, \mathbf{x}, y\right)$ denotes its gradient with respect to $\mathbf{x}$.
% \\footnote{***Define $\delta$ here}
% FGSM initializes $\delta$ to zero and uses $\epsilon$ as attack step size so that it generates the adversarial example by one step as $\mathbf{x}^{\prime}=\mathbf{x} + \epsilon\cdot\mathrm{sign}(\nabla_\mathbf{x} \ell\left(\bm{\theta}, \mathbf{x}, y\right))$.

\vspace{-0.35cm}
\paragraph{Projected Gradient Descent (PGD)}
Instead of generating one-step perturbations as in FGSM, Kurakin \etal~\cite{pgd17} propose the PGD method by applying a small number of iterative steps. To ensure the perturbation is in $\epsilon$-neighborhood of the original image, the PGD method clips the intermediate results after each iteration as
{\small
\begin{equation*}
\bm{\delta}_{t+1}=\mathrm{clip}_{ \epsilon}(\bm{\delta}_{t}+\xi\cdot\mathrm{sign}(\nabla_\mathbf{x} \ell\left(\bm{\theta}, \mathbf{x}+\bm{\delta}_{t}, y\right))),
\end{equation*}}
where $\bm{\delta}_{t}$ is the perturbation generated in the $t$-th steps, $\xi$ is the attack step size. $\mathrm{clip}_{\epsilon}(\cdot)$ means to elementwisely clip the input to lie within an interval $[-\epsilon, \epsilon]$, \ie $\|\bm{\delta}_{t+1}\|_{\infty}\leq\epsilon$. % which is equivalent to project the perturbation onto the $\epsilon$-ball around the origin.
% $x_i^\prime$ is the intermediate result of adversarial example and $x_{i+1}^\prime$ is the $i+1$ step of the iterations. The iterative process stops when $x_{i+1}^\prime$ is misclassified by the model.
% The PGD attack starts from a random perturbation in $L_n$\footnote{***What is $n$?} ball around a sample and projects perturbation back into $\epsilon$-ball.

\vspace{-0.35cm}
\paragraph{Adversarial Training}
Adversarial training is an effective method for defending adversarial attacks \cite{fgsm15, pgd17, pgd7, fast_fgsm}. %Deep neural networks can represent functions that resist adversarial perturbation.
%Trained with adversarial loss functions, deep neural networks are more robust to adversarial attacks.
Goodfellow \etal~\cite{fgsm15} %use an adversarial objective function to perform the adversarial training in the training stage of the model. It
leverage the FGSM as a regularizer to train deep neural networks and make the model more resistant to adversarial examples. Wong \etal~\cite{fast_fgsm} use FGSM adversarial training with random initialization
for the perturbation. The proposed method can speed up the adversarial training process and it is as effective as the PGD-based adversarial training.

\subsection{Multi-Objective Optimization}

Multi-objective optimization aims to optimize more than one objective function simultaneously. Among different techniques to solve multi-objective problems, we are interested in gradient-based multi-objective optimization algorithms \cite{desideri12, fliege2000steepest, schaffler2002stochastic}, which leverage the Karush-Kuhn-Tucker (KKT) conditions \cite{kuhn2014nonlinear} to find a common descent direction for all objectives.

In this paper, we utilize one such method, \ie MGDA \cite{desideri12}. %, to make a trade-off between the performance,  resource constraint, and robustness.
With $n$ objective functions $\{\mathcal{L}_i(\bm{\theta})\}_{i=1}^n$ to be minimized, MGDA is an iterative method by first solving the following quadratic programming problem as
{\small
\begin{equation}
\begin{aligned}
\min_{\gamma_1,\cdots,\gamma_n} \ \left\|\sum_{i=1}^n\gamma_i \nabla_{\bm{\theta}}\mathcal{L}_i(\bm{\theta}) \right\|^2_2 \
\mathrm{s.t.}\ \gamma_i \geq 0,\ \sum_{i=1}^n \gamma_i=1,
\end{aligned}
\label{eq:mgda}
\end{equation}}
where $\|\cdot\|_2$ denotes the $\ell_2$ norm of a vector and $\gamma_i$ can be viewed as a weight for the $i$th objective, and then minimizing $\sum_{i=1}^n\gamma_i\mathcal{L}_i(\bm{\theta})$ with respect to $\bm{\theta}$. When convergent, the MGDA can find a Pareto-stationary solution. %Obviously, problem (\ref{eq:mgda}) is a quadratic programming problem. %, leading to huge computational complexity especially in large-scale learning problems.
%Sener and Koltun \cite{sk18} used the Frank-Wolfe algorithm to solve this problem and apply it to multi-task deep learning problems.

\subsection{Multi-Objective NAS}

%NAS is becoming popular because it does not need human experts to design the architecture and it can also find better architectures than human did.
Most NAS methods focus on searching architectures with the best accuracy. However, in real-world applications, other factors, such as model size and robustness, must be considered. To take those factors into consideration, some works on multi-objective NAS have been proposed in recent years. LEMONADE \cite{emh19} considers two objectives, including maximizing the validation accuracy and minimizing the number of parameters. It is based on the evolutionary algorithm and thus the search cost is quite high. MnasNet \cite{tcpvshl19} uses a reinforcement learning approach to optimize both the accuracy and inference latency together when searching the architecture. %It does not use the commonly used FLOPS to measure the latency because the FLOPS does not always reflect the actual latency. It directly measures real-world inference latency by deploying the model on mobile phones. MnasNet used a reinforcement learning approach to find Pareto optimal points for the multi-objective optimization problem.
Chen~\etal \cite{cztl20} perform the neural architecture search based on the reinforcement learning to optimize three objectives, including the maximization of the validation accuracy, the minimization of the number of parameters, and the minimization of the number of FLOPs.  %The proposed method can approximate the full Pareto front accurately and efficiently.
FBNet \cite{wu2019fbnet} also considers both the accuracy and model latency when searching the architecture via a gradient-based method to solve the corresponding multi-objective problem. %It also uses real-world latency. As the search cost of the reinforcement learning method proposed in MnasNet is quite high, FBNet uses a gradient-based method for the multi-objective problem. Therefore, the search cost of FBNet is much smaller than MnasNet.
Built on DARTS, RC-DARTS \cite{jin2019rc} considers to search architectures with high accuracy while constraining model parameters of the searched architecture below a threshold. %The model parameter is considered as the constraint in the process of search.
Therefore, the proposed objective function is formulated as a constrained optimization problem and a projected gradient descent method is proposed to solve it. Based on DARTS, GOLD-NAS \cite{bi2020gold} considers three objectives: maximizing the validation accuracy, minimizing the number of parameters, and minimizing the number of FLOPs, and by enlarging the search space, it proposes a one-level optimization algorithm instead of the bi-level optimization.

\section{The E2RNAS Method}

In this section, we present the proposed E2RNAS method. We first give an overview of the DARTS method and then introduce how to achieve the robustness and formulate the objective to constrain the number of parameters in the search architecture. Finally we present the bi-level multi-objective problem of the proposed E2RNAS method as well as its optimization.

\subsection{Preliminary: DARTS}

%In this section, we first retrospect the baseline DARTS \cite{lsy19} and clearly claim the notations in this paper.
DARTS \cite{lsy19} aims to learn a Directed Acyclic Graph (DAG) called cell, which can be stacked to form a neural network architecture. Each cell consists of $N$ nodes $\{x_i\}_{i=0}^{N-1}$, each of which denotes a hidden representation. $\mathcal{O}$ denotes a discrete operation space. The edge $(x_i, x_j)$ of the DAG represents an operation function $o(\cdot)$ (\eg skip connection or $3\times 3$ pooling) from $\mathcal{O}$ with a probability $\alpha^{(i,j)}_o$ to perform at the node $x_i$. Therefore, we can formulate each edge $(x_i, x_j)$ as a weighted sum function to combine all the operations in $\mathcal{O}$ as $f_{i,j}(x_i) = \sum_{o\in \mathcal{O}}\frac{\exp(\alpha^{(i,j)}_o)}{\sum_{o^{\prime}\in\mathcal{O}}\exp(\alpha^{(i,j)}_{o^{\prime}})}o(x_i)$.
% {\small
% \begin{equation*}
%     f_{i,j}(x_i) = \sum_{o\in \mathcal{O}}\frac{\exp(\alpha^{(i,j)}_o)}{\sum_{o^{\prime}\in\mathcal{O}}\exp(\alpha^{(i,j)}_{o^{\prime}})}o(x_i).
% \end{equation*}
% }
An intermediate node $x_j$ is the sum of its predecessors, \ie $x_j=\sum_{i<j}f_{i,j}(x_i)$. The output of the cell, \ie node $x_{N-1}$, is the concatenation of all the output of nodes excluding the two input nodes $x_0$ and $x_1$. Therefore, $\bm{\alpha}=\{\alpha^{(i,j)}_o\}_{(i,j)\in\mathbf{E},~o\in\mathcal{O}}$ can parameterize the searched architecture, where $\mathbf{E}$ denotes the set of all the edges from all the cells.

Let $\mathbf{X}_{tr}$ denote the training dataset and $\mathbf{Y}_{tr}$ denote the corresponding set of labels. Similarly, the validation dataset and labels are denoted by $\mathbf{X}_{val}$ and $\mathbf{Y}_{val}$. We use $\bm{\theta}$ to denote all the weights of the neural network and  $\ell(\bm{\theta},\mathbf{x},y)$ to denote the loss function. DARTS is to solve a bi-level optimization problem as
\begin{equation}
\begin{aligned}
\min_{\bm{\alpha}} \quad & \mathcal{L}_{val}(\bm{\theta}^*(\bm{\alpha}),\bm{\alpha})\\
\mathrm{s.t.} \quad & \bm{\theta}^*(\bm{\alpha}) = {\arg\min}_{\bm{\theta}}~\mathcal{L}_{tr}(\bm{\theta},\bm{\alpha}),
\end{aligned}
\label{eq:darts_loss}
\end{equation}
where $\mathcal{L}_{tr}(\bm{\theta},\bm{\alpha})=\frac{1}{|\mathbf{X}_{tr}|}\sum_{(\mathbf{x},y)\in(\mathbf{X}_{tr},\mathbf{Y}_{tr})}~\ell(\bm{\theta},\mathbf{x},y)$ and $\mathcal{L}_{val}(\bm{\theta},\bm{\alpha})=\frac{1}{|\mathbf{X}_{val}|}\sum_{(\mathbf{x},y)\in(\mathbf{X}_{val},\mathbf{Y}_{val})}~\ell(\bm{\theta},\mathbf{x},y)$ represent the training and validation losses, respectively. Here $\min_{\bm{\alpha}} \mathcal{L}_{val}(\bm{\theta}^*(\bm{\alpha}),\bm{\alpha})$ is called the \emph{upper-level problem} and $\min_{\bm{\theta}}\mathcal{L}_{tr}(\bm{\theta},\bm{\alpha})$ is called the \emph{lower-level problem}.

When the search procedure finishes, the final architecture can be determined by the operation with the largest probability in each cell, \ie $o^{(i,j)}={\arg\max}_{o\in\mathcal{O}}~~\alpha^{(i,j)}_o$.

\subsection{Adversarial Training for Robustness}

In E2RNAS, we expect the searched architecture to be robust, which means that for the trained model with the searched architecture, its performance is stable when adding some perturbation to the dataset. To improve the robustness of the searched architecture, we leverage the adversarial training method in \cite{fast_fgsm} to train a robust model.
% we make two adjustments to the bi-level optimization in DARTS.

% Secondly, we evaluate the robustness of validation dataset in the outer loop and formulate as an objective function to optimize with other objectives. More details are in Section \ref{sec:bi-level for E2RNAS}.

Following \cite{fast_fgsm}, for each sample $\mathbf{x}$ and its corresponding label $y$, we can generate a perturbation for $\mathbf{x}$ using one single step as
\begin{equation*}
\bm{\delta}^{\prime}(\mathbf{x}, y) = \mathrm{clip}_{ \epsilon}(\bm{\delta}+\xi\cdot\mathrm{sign}(\nabla_{\mathbf{x}} \ell(\bm{\theta},\mathbf{x}+\bm{\delta},y))),
% \label{eq:adv_sample}
\end{equation*}
where $\epsilon$ is the perturbation size, $\bm{\delta}$ is randomly initialized with an uniform distribution on the interval $[-\epsilon, \epsilon]$, and $\xi$ is the attack step size. %$\nabla_{\mathbf{x}}\ell(\bm{\theta},\mathbf{x}+\delta,y)$ is the derivative of loss function $\ell(\bm{\theta},\mathbf{x}+\delta,y)$ with respect to $\mathbf{x}$, $\mathrm{sign}(\cdot)$ denotes the sign function, $\mathrm{clip}_{\epsilon}(\cdot)$ means to clip the input to lie within the range $(-\epsilon, \epsilon)$.
Therefore, we generate the adversarial instance as $\mathbf{x}^{\prime} = \mathbf{x} + \bm{\delta}^{\prime}(\mathbf{x}, y)$. Obviously, the FGSM is a special case of this method when $\bm{\delta}$ is initialized with zero and $\xi=\epsilon$. This FGSM-based adversarial training method with random initialization for $\bm{\delta}$ \cite{fast_fgsm} can effectively defense the PGD adversarial attack \cite{pgd17}, while not adding much computational cost in the architecture search procedure.

We use these perturbed data to learn the network parameters $\bm{\theta}$ so that the trained model can defense adversarial attacks. Therefore, we aim to minimize the training loss of the perturbed data as
\begin{equation}
\mathcal{L}_{tr}^{\mathrm{adv}}(\bm{\theta},\bm{\alpha}) = \frac{1}{|\mathbf{X}_{tr}|} \sum_{(\mathbf{x},y)\in(\mathbf{X}_{tr},\mathbf{Y}_{tr})}\ell(\bm{\theta},\mathbf{x}^{\prime},y).
\label{eq:adv_train}
\end{equation}
Note that this adversarial training method trains the model only on adversarial examples, which is different from the FGSM-based adversarial training method \cite{fgsm15} that uses them as a regularization term for training.

% Similarly to Eq. (\ref{eq:adv_train}), we can formulate the objective function of robustness $\mathcal{L}_{val}^{\mathrm{adv}}(\bm{\theta},\bm{\alpha})$ in the outer loop to evaluate the robustness of searched architecture.

\subsection{Objective Function of Resource Constraints}

% Resource constraints are one of the important fa ctors we consider.
Architectures with a small number of parameters have more application scenarios even in resource-constrained mobile devices. Therefore, we regard resource constraints as one of the desired objectives.

By following DARTS \cite{lsy19}, we determine the operation of each cell in the final architecture as the one with the largest probabilities. So the number of parameters in an architecture can be computed as
\begin{equation}
    \mathcal{N}(\bm{\alpha}) = \sum_{(i,j)\in\mathbf{E}} n_{{\arg\max}_{o\in\mathcal{O}}~\alpha^{(i,j)}_o},
\label{eq:num_param}
\end{equation}
where $n_o$ denotes the number of parameters corresponding to the operation $o$.
% where $L$ denotes the maximum number of layers in the architecture, $n^i_j$ denotes the number of parameters contained in the $j$th candidate cell for the $i$th layer with $\alpha^i_j$ representing the corresponding probability to be selected.

Note that ${\arg\max}_{o\in\mathcal{O}}~\alpha^{(i,j)}_o$ in Eq.~(\ref{eq:num_param}) is a non-differentiable operation, making the computation of the gradient of $\mathcal{N}(\bm{\alpha})$ with respect to $\bm{\alpha}$ infeasible. To make such operation differentiable, we use the softmax trick to approximate the $\arg\max$ operation and then formulate the approximation $\hat{\mathcal{N}}(\bm{\alpha})$ as
\begin{equation}
    \hat{\mathcal{N}}(\bm{\alpha}) = \sum_{(i,j)\in\mathbf{E}} \sum_{o\in\mathcal{O}}\frac{\exp(\alpha^{(i,j)}_o)}{\sum_{o^{\prime}\in\mathcal{O}}\exp(\alpha^{(i,j)}_{o^{\prime}})} n_o.
\label{eq:num_param_soft}
\end{equation}
% where $\mathbf{E}$ is a set of all the edges from all cells, and $n_o$ represents the parameter size of the operation $o$.

Furthermore, to prevent the model to search over-simplified architectures (\ie the one containing too many parameter-free operations) that leads to unsatisfactory performance, we add a lower bound $L$ to the parameter size in Eq. (\ref{eq:num_param_soft}), \ie $\hat{\mathcal{N}}(\bm{\alpha})\geq L$. Therefore, the objective function of the resource constraint can be formulated as
\begin{equation}
    \Psi(\bm{\alpha}) = \max(\hat{\mathcal{N}}(\bm{\alpha}), L).
%  \Psi(\bm{\alpha}) = \mathcal{N}(\bm{\alpha})+\max(0, \mathcal{N}(\bm{\alpha})-U, L-\mathcal{N}(\bm{\alpha})).
    \label{eq:num_param_loss}
\end{equation}
Different from RC-DARTS \cite{jin2019rc} that directly adds the resource constraint into the original DARTS objective function (\ref{eq:darts_loss}) as a constraint and formulates the objective function as a constrained optimization problem, here we take it as an objective function.

\subsection{Bi-level Multi-Objective Formulation} \label{sec:bi-level for E2RNAS}

E2RNAS aims to search the architecture parameter $\bm{\alpha}$ to minimize the validation loss for the effectiveness and the number of parameters for the efficiency, while achieving the robustness via the adversarial training.
% where the robustness is encoded in the cost function on the training dataset and doesn't explicitly formulate as a objective function with respect to $\bm{\alpha}$.
%The training dataset is used to learn a robust model with its weight $\bm{\theta}$ via adversarial training.
Thus, we combine Eqs. (\ref{eq:adv_train}) and (\ref{eq:num_param_loss}) as well as the adversarial training to formulate the entire objective function as
\begin{equation}
\begin{aligned}
\min_{\bm{\alpha}} ~~& (\mathcal{L}_{val}(\bm{\theta}^*(\bm{\alpha}),\bm{\alpha}), \Psi(\bm{\alpha}))\\
\mathrm{s.t.} ~~& \bm{\theta}^*(\bm{\alpha}) = {\arg\min}_{\bm{\theta}}~\mathcal{L}^\mathrm{adv}_{tr}(\bm{\theta},\bm{\alpha}).
\label{eq:obj_fun}
\end{aligned}
\end{equation}
Problem (\ref{eq:obj_fun}) is similar to the bi-level optimization problem (\ref{eq:darts_loss}) in the DARTS, where the \emph{lower-level problem} (\ie $\min_{\bm{\theta}}~\mathcal{L}^\mathrm{adv}_{tr}(\bm{\theta},\bm{\alpha})$) is similar, but there exists significant differences in that the \emph{upper-level problem} (\ie $\min_{\bm{\alpha}} (\mathcal{L}_{val}(\bm{\theta}^*(\bm{\alpha}),\bm{\alpha}), \Psi(\bm{\alpha}))$) contains two objectives. So problem \eqref{eq:obj_fun} is a bi-level multi-objective optimization problem which is a generalization of problem \eqref{eq:darts_loss} in the DARTS. There are few works on bi-level multi-objective optimization \cite{calvete2010linear, deb2009solving, zhang2012improved, ruuska2012constructing} and to the best of our knowledge, the proposed optimization algorithm as introduced in the following is the first gradient-based algorithm to solve general bi-level multi-objective  optimization problems.

Problem (\ref{eq:obj_fun}) can be understood as a two-stage optimization. Firstly, when given an architecture parameter $\bm{\alpha}$, we can learn a robust model with optimal model weights $\bm{\theta}^{*}$ via the empirical risk minimization on adversarial examples. Secondly, given  $\bm{\theta}^*$, the architecture parameter $\bm{\alpha}$ is updated on the validation dataset by making a trade-off between its performance and model size. Therefore, we can solve problem (\ref{eq:obj_fun}) in two stages, which are described as follows.

\vspace{-0.3cm}
\paragraph{Updating $\bm{\theta}$}
Given the architecture parameter $\bm{\alpha}_t$, %(\ie the architecture of the model is determined),
$\bm{\theta}$ can be simply updated as
\begin{equation}
    \bm{\theta}_{t+1} = \bm{\theta}_{t} - \eta_{\bm{\theta}}\nabla_{\bm{\theta}} \mathcal{L}_{tr}^{\mathrm{adv}}(\bm{\theta}_{t},\bm{\alpha}_{t}),
\label{eq:theta_update}
\end{equation}
where $t$ denotes the index of the iteration and $\eta_{\bm{\theta}}$ denotes the learning rate.

\vspace{-0.3cm}
\paragraph{Updating $\bm{\alpha}$} After obtaining $\bm{\theta}_{t+1}$, we can optimize the upper-level problem to update the architecture parameter $\bm{\alpha}$. As the upper-level problem is a multi-objective optimization problem, we adopt the MGDA to solve it. In MGDA, we first need to solve problem (\ref{eq:mgda}), which requires the computation of the gradients of the two objectives with respect to $\bm{\alpha}$. The gradient of $\Psi(\bm{\alpha}_t)$ with respect to $\bm{\alpha}$ is easy to compute, while the gradient of $\mathcal{L}_{val}(\bm{\theta}^*(\bm{\alpha}_t),\bm{\alpha}_t)$ %and $\mathcal{L}_{val}^{\mathrm{adv}}(\bm{\theta}^*(\bm{\alpha}_t),\bm{\alpha}_t)$
with respect to $\bm{\alpha}$ is a bit complicate as $\bm{\theta}^*(\bm{\alpha}_t)$ is also a function of $\bm{\alpha}$ and it is too expensive to obtain $\bm{\theta}^*(\bm{\alpha}_t)$. Therefore, we use a second-order approximation as
\begin{equation}
\begin{aligned}
    &\nabla_{\bm{\alpha}}\mathcal{L}_{val}(\bm{\theta}^*(\bm{\alpha}_t),\bm{\alpha}_t)\\
    \approx & \nabla_{\bm{\alpha}}\mathcal{L}_{val}(\bm{\theta}_{t+1}-\eta_{\bm{\theta}}\nabla_{\bm{\theta}}\mathcal{L}_{tr}^{\mathrm{adv}}(\bm{\theta}_{t+1},\bm{\alpha}_t),\bm{\alpha}_t),
\label{eq:second_order}
\end{aligned}
\end{equation}
% where $\xi$ denotes the learning rate for inner optimization.
Obviously when $\eta_{\bm{\theta}}=0$, $\bm{\theta}_{t+1}$ becomes an approximation of $\bm{\theta}^*(\bm{\alpha}_t)$ and Eq. (\ref{eq:second_order}) degenerates to the first-order approximation, which can speed up the gradient computation and reduce the memory cost but lead to worse performance \cite{lsy19}. So we use the second-order approximation in Eq. (\ref{eq:second_order}).
% Similarly, the gradient of $\mathcal{L}_{val}^{\mathrm{adv}}(\bm{\theta}^*(\bm{\alpha}_t),\bm{\alpha}_t)$ can be computed approximately.
Then due to the two objectives in the upper-level problem of problem (\ref{eq:obj_fun}), we can simplify problem (\ref{eq:mgda}) as a one-dimensional quadratic function of $\gamma$ as
\begin{equation}
    \min_{0\le\gamma\le1} \left\|\gamma\bm{u}_1+(1-\gamma)\bm{u}_2\right\|^2_2,
    % \mathrm{s.t.}\quad &n=1,
\label{eq:subproblem}
\end{equation}
where $\bm{u}_1=\nabla_{\bm{\alpha}_t}\mathcal{L}_{val}(\bm{\theta}^*(\bm{\alpha}_t),\bm{\alpha}_t)$ and $\bm{u}_2=\nabla_{\bm{\alpha}_t}\Psi(\bm{\alpha}_t)$ denote the gradients of two objectives, respectively. Here $\gamma$ can be viewed the weight for the first objective and $1-\gamma$ is for the second objective. It is easy to show that problem \eqref{eq:subproblem} has an analytical solution as
\begin{equation}
\hat{\gamma} = \max\left(\min\left(\frac{(\bm{u}_2-\bm{u}_1)^T \bm{u}_2}{\|\bm{u}_1-\bm{u}_2\|_2^2}, 1\right), 0\right).
\label{eq:gamma}
\end{equation}
% When the number of the objective function exceed three, the problem (\ref{eq:mgda}) has not the analytical solutions and Sener and Koltun \cite{sk18} applied the Frank-Wolfe algorithm to solve it based on analytically solving each subproblem (\ref{eq:subproblem}) for the line search. In this paper, we use their method to solve for a Pareto stationary solution $\bm{\hat{\gamma}}=\{\hat{\gamma}_1, \hat{\gamma}_2, \hat{\gamma}_3\}$. More details are in Algorithm 2 in \cite{sk18}.
After that, we can update $\bm{\alpha}_t$ by minimizing $U(\bm{\theta}^*(\bm{\alpha}_t),\bm{\alpha}_t) = \hat{\gamma}\mathcal{L}_{val}(\bm{\theta}^*(\bm{\alpha}_t),\bm{\alpha}_t)+(1-\hat{\gamma})\Psi(\bm{\alpha}_t)$ as
\begin{equation}
\hskip -0.05in\bm{\alpha}_{t+1} = \bm{\alpha}_t - \eta_{\bm{\alpha}}\nabla_{\bm{\alpha}_t}U(\bm{\theta}^*(\bm{\alpha}_t),\bm{\alpha}_t),
\label{eq:alpha_update}
\end{equation}
where $\eta_{\bm{\alpha}}$ denotes the learning rate for $\bm{\alpha}$.

%In practice, we can apply automatic differentiation techniques based on popular deep learning framework like PyTorch to easily compute the gradient of two objective function $\mathcal{L}_{val}(\bm{\theta}^*(\bm{\alpha}),\bm{\alpha})$
% , $\mathcal{L}_{val}^{\mathrm{adv}}(\bm{\theta}^*(\bm{\alpha}),\bm{\alpha})$
%and $\Psi(\bm{\alpha})$.
% \vspace{-0.4cm}

\vspace{-0.4cm}
\begin{figure}[!htbp]
	\renewcommand{\algorithmicrequire}{\textbf{Input:}}
	\renewcommand{\algorithmicensure}{\textbf{Output:}}
%	\removelatexerror
	\begin{algorithm}[H]
	\caption{E2RNAS}
	\label{alg:E2RNAS}
		\begin{algorithmic}[1]
			\REQUIRE Dataset $\{\mathbf{X}_{tr},\mathbf{Y}_{tr}, \mathbf{X}_{val},\mathbf{Y}_{val}\}$, batch size $B$, perturbation size $\epsilon$, minimum constraint $L$, learning rates $\eta_{\bm{\alpha}}$ and $\eta_{\bm{\theta}}$
			\ENSURE Learned architecture parameter $\bm{\alpha}$
			\STATE Randomly initialized $\bm{\alpha}_0$ and $\bm{\theta}_0$;
			\STATE $t:=0$;
			\WHILE{not converged}
			\STATE Sample a mini-batch of size $B$;
% 			\STATE Generate the adversarial samples using Eq. (\ref{eq:adv_sample}).
			\STATE Compute $\mathcal{L}_{tr}^{\mathrm{adv}}(\bm{\theta}_t,\bm{\alpha}_t)$ according to Eq. (\ref{eq:adv_train});  %and resource constraints loss $\Psi(\bm{\alpha})$ using Eq. (\ref{eq:adv_train}) and Eq. (\ref{eq:num_param_loss}), respectively.
			\STATE Update $\bm{\theta}_{t+1}$ according to Eq. (\ref{eq:theta_update});
			\STATE Compute two objective functions $\mathcal{L}_{val}(\bm{\theta}^*(\bm{\alpha}_t),\bm{\alpha}_t)$, $\Psi(\bm{\alpha}_t)$ and the corresponding gradients;
			\STATE Compute $\hat{\gamma}$ according to Eq. (\ref{eq:gamma});
			\STATE Update $\bm{\alpha}_t$ according to Eq. (\ref{eq:alpha_update});
			\STATE $t:=t+1$;
			\ENDWHILE
		\end{algorithmic}
	\end{algorithm}
\end{figure}
\vspace{-1cm}

\paragraph{Comparison between E2RNAS and DARTS}
Though the proposed E2RNAS method is based on the DARTS, there are two key differences between them, which are shown in Figure \ref{fig:motivation}. Firstly, E2RNAS adopts the adversarial training to improve the robustness of the corresponding neural network. Secondly, E2RNAS evaluates model with two objectives: minimizing the validation loss for the effectiveness and the number of parameters for the efficiency. Therefore, E2RNAS can search an effective, efficient, and robust architecture. The whole algorithm is summarized in Algorithm \ref{alg:E2RNAS}.

\section{Experiments}

In this section, we empirically evaluate the proposed E2RNAS method on three image datasets, including CIFAR-10 \cite{krizhevsky2009learning}, CIFAR-100 \cite{krizhevsky2009learning}, and SVHN \cite{netzer2011reading}. Details about these datasets are presented in the Appendix.
%All the implementation details are in the Appendix.
% \subsection{Dataset}
% Experiments are conducted on three image datasets, including CIFAR-10 \cite{krizhevsky2009learning}, CIFAR-100 \cite{krizhevsky2009learning}, and SVHN \cite{netzer2011reading}. %The CIFAR-10 dataset is the most commonly used dataset in neural architecture search research.
% The CIFAR-10 dataset contains 50,000 training images and 10,000 testing images from 10 classes, each of which has 6,000 images with a $32\times 32$ resolution in total.  %The resolution of the images is $32\times 32$.
% The CIFAR-100 dataset contains 100 classes, which are grouped into 20 superclasses, with 500 training images and 100 testing images for each class. The Street View House Numbers (SVHN) dataset consists of the images of house numbers captured from the Google street view and  %The dataset is designed for machine learning and object recognition research. It has a minimal requirement on data preprocessing and formatting.
% it contains 10 classes with 73,257 images used for training and 26,032 image for testing. %Additional 531,131 digits, which are somewhat less difficult samples, are used as extra training data.

\subsection{Implementation Details}
\paragraph{Search Space}
The search space adopts the same setting as DARTS \cite{lsy19}. There are two types of cells, \ie the reduction cell and the normal cell. The reduction cell is located at the $1/3$ and $2/3$ of the total depth of the network and other cells belong to the normal cell. For both reduction and normal cells, there are $7$ nodes in each cell, including four intermediate nodes, two input nodes, and one output node. In both normal and reduction cell, the set of operations $\mathcal{O}$ contains eight operations, including $3\times3$ separable convolutions, $5\times5$ separable convolutions, $3\times3$ dilated separable convolutions, $5\times5$ dilated separable convolutions, $3\times3$ max pooling, $3\times3$ average pooling, identity, zero. For the convolution operator, the ReLU-Conv-BN order is used.

\vspace{-0.3cm}
\paragraph{Training Settings}
By following DARTS \cite{lsy19}, a half of the standard training set is used for training a model and the other half for validation. A small network of 8 cells is trained via the FGSM-based adversarial training method \cite{fast_fgsm} in Eq. (\ref{eq:adv_train}) with the batch size as 64 and initial channels as 16 for 50 epochs. Following the setting of \cite{fast_fgsm}, the perturbation of the FGSM adversary is randomly initialized from the uniform distribution in $[-\epsilon,\epsilon]$, where $\epsilon = 2/255$. The attack step size $\xi$ is set to $1.25\epsilon$.
The SGD optimizer with the momentum $0.9$ and the weight decay $3\times 10^{-4}$ is used. The proposed method is implemented in PyTorch 0.3.1 and all the experiments are conducted in Tesla V100S GPUs with 32G CUDA memory.

\vspace{-0.3cm}
\paragraph{Evaluation Settings}
A large network of 20 cells is trained on the full training set for 600 epochs, with the batch size as 96, the initial number of channels 36, a cutout of length 16, the dropout probability 0.2, and auxiliary towers of weight 0.4. To make the model size comparable, we adjust the initial channels of each cell for both DARTS and the proposed E2RNAS method, which is denoted by ``$\{\text{model-C}\#\text{channels}\}$". The accuracy is tested on the full testing set. Adversarial examples are generated using the PGD attack \cite{pgd17} with the perturbation size $\epsilon = 2/255$ on the testing set. The PGD attack takes 10 iterative steps with the step size of $2.5\epsilon$ as suggested in \cite{pgd7}.

\begin{table*}[!htbp]
\centering
\begin{tabular}{lccccc}
\toprule
\multirow{2}*{\textbf{Architecture}} & \textbf{Test Err.}  & \textbf{Params} & \textbf{PGD Acc.} & \textbf{Search Cost}& \multirow{2}*{\textbf{Search Method}} \\
~ & \textbf{(\%)}~$\downarrow$ & \textbf{(MB)}~$\downarrow$ & \textbf{(\%)}~$\uparrow$ & \textbf{(GPU days)}~$\downarrow$ & ~ \\
\midrule
DenseNet-BC$^\dag$ \cite{huang2017densely} & 3.46 & 25.6 &  - & - & manual \\
\midrule
NASNet-A \cite{zoph2016neural} & 2.65 & 3.3 & - & 2000 & RL \\
AmoebaNet-B \cite{real2019regularized}& 2.55$\pm$0.05 & 2.8 & - &3150& evolution \\
Hireachical Evolution$^\dag$ \cite{liu2017hierarchical} & 3.75$\pm$0.12 & 15.7 & - &300& evolution \\
PNAS$^\dag$ \cite{liu2018progressive} & 3.41$\pm$0.09 & 3.2 & - &225& SMBO \\
ENAS \cite{pham2018efficient} & 2.89 & 4.6 & -&0.5 & RL \\
% NAONet$^\dag$ \cite{luo2018neural} & 3.53 & 3.1 & - & NAO \\
\midrule
DARTS$^\ddag$ \cite{lsy19} &2.59  &3.349  &6.57 &0.595$^{*}$& gradient-based\\
DARTS-C28$^\ddag$    &2.68  &2.061  &5.42 &0.595$^{*}$& gradient-based\\
DARTS-C20$^\ddag$    &3.15  &1.083  &3.90 &0.595$^{*}$& gradient-based\\
DARTS-C12$^\ddag$    &3.09  &0.416  &3.08 &0.595$^{*}$& gradient-based\\
P-DARTS$^\ddag$ \cite{chen2019progressive}   &2.59  &3.434  &8.35 &0.247$^{*}$& gradient-based\\
PC-DARTS$^\ddag$ \cite{xu2019pc}    &2.65  &3.635  &9.53 &0.426$^{*}$& gradient-based\\
\midrule
E2RNAS-C46    &3.64  &3.383  &10.21 &0.836$^{*}$& gradient-based\\
E2RNAS-C36    &4.19  &2.102  &9.61 &0.836$^{*}$& gradient-based\\
E2RNAS-C25    &4.86  &1.042  &7.76 &0.836$^{*}$& gradient-based\\
E2RNAS-C16    &6.03  &0.449  &6.76 &0.836$^{*}$& gradient-based\\
\bottomrule
\end{tabular}
\vskip 0.05in
\caption{Comparison with state-of-the-art NAS methods on the CIFAR-10 dataset. $\dag$ represents training without the cutout augmentation. $\ddag$ indicates the use of the  code released by original authors. $\uparrow$ indicates a larger value is better, while $\downarrow$ indicates a lower value is better. ``$\{\text{model-C}\#\text{channels}\}$" means the architecture searched by ``model" is evaluated with the initial number of channels as ``channels". $*$ means the search cost is recorded on a single Tesla V100S GPU.}
\label{tab:result_cifar10}
\end{table*}

\subsection{Analysis on Experimental Results}
% \paragraph{ Settings.}
% We employ our adversarial training loss and number of parameter objective (AP) on DARTS and its popular variants to search CNN cells on CIFAR-10.

% \paragraph{ Results.}  Table ~\ref{tab:result} summarizes the comparision of our methods (AP) with the state-of-the-art algorithms.

\begin{figure}[!htbp]
  \includegraphics[width=\linewidth]{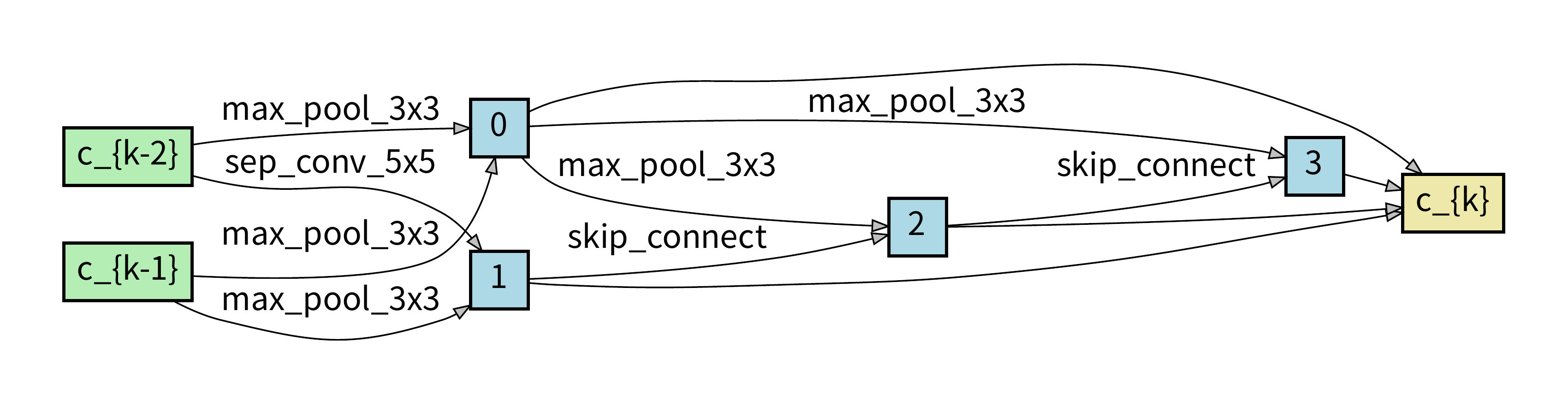}
  \caption{The normal cell in E2RNAS learned on CIFAR-10.}
  \label{fig:normal}
\end{figure}

\begin{figure}[!htbp]
  \includegraphics[width=\linewidth]{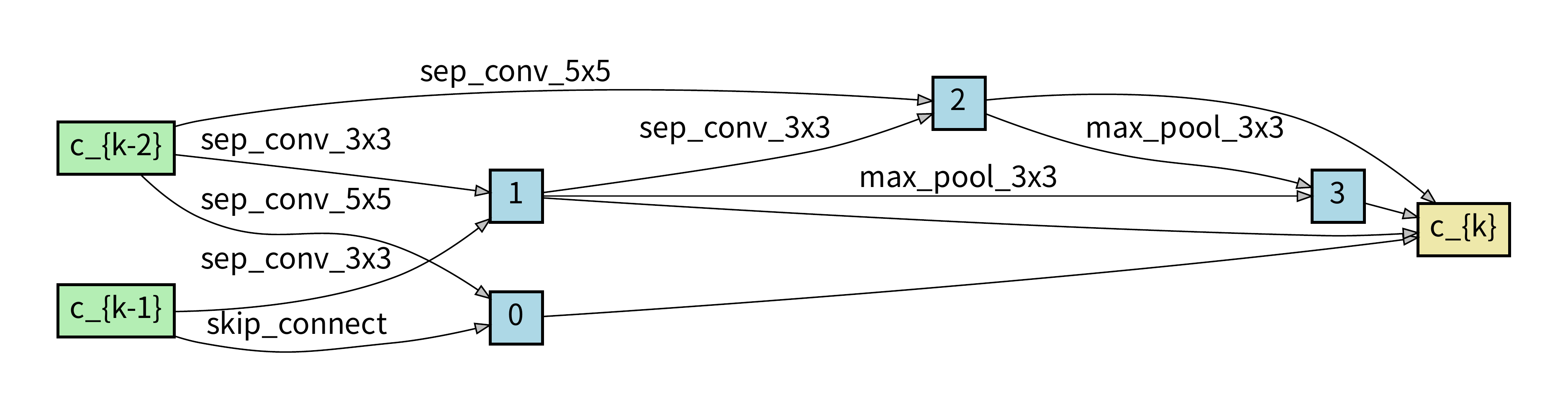}
  \caption{The reduction cell in E2RNAS learned on CIFAR-10.}
  \label{fig:reduction}
\end{figure}

\paragraph{Search Architecture on CIFAR-10}
The normal and reduction cells searched by the E2RNAS method on the CIFAR-10 dataset are presented in Figures \ref{fig:normal} and \ref{fig:reduction}, respectively. Different from DARTS \cite{lsy19}, the reduction cell in E2RNAS contains many convolution operations and the normal cell only includes one operation with parameters (\ie the $5\times 5$ separable convolution). Thus, the parameter size of the architecture searched by E2RNAS is lower than that of DARTS because E2RNAS searched an architecture with fewer reduction cells.

\vspace{-0.3cm}
\paragraph{Architecture Evaluation on CIFAR-10}
The comparison of the proposed E2RNAS method with state-of-the-art NAS methods on the CIFAR-10 dataset is shown in Table \ref{tab:result_cifar10}. Notably, E2RNAS outperforms these NAS methods in \cite{zoph2016neural, real2019regularized, liu2017hierarchical, liu2018progressive} by searching for a more lightweight architecture with lower search costs of three to four orders of magnitude and a slightly higher test error rate. Moreover, although ENAS \cite{pham2018efficient} slightly outperforms E2RNAS in the test accuracy and search time, it finds a deeper architecture with about doubled model size (\ie 4.6MB for ``ENAS" \vs 2.102MB for ``E2RNAS-C36").

\begin{table}[!htbp]
\centering
\resizebox{\linewidth}{!}{
\begin{tabular}{clccc}
\toprule
\multirow{2}*{\textbf{Dataset}} & \multirow{2}*{\textbf{Architecture}} & \textbf{Test Err.}  & \textbf{Params} & \textbf{PGD Acc.} \\
~ & ~ & \textbf{(\%)}~$\downarrow$ & \textbf{(MB)}~$\downarrow$ & \textbf{(\%)}~$\uparrow$\\
\midrule
\multirow{10}*{\rotatebox{90}{\textbf{CIFAR-100}}} & DARTS$^\ddag$ \cite{lsy19}   &17.17 &3.401   & 2.06 \\
~ & DARTS-C34$^\ddag$ & 17.70 & 3.047 & 1.67 \\
~ & DARTS-C27$^\ddag$ & 17.78 & 1.960 & 1.70 \\
~ & DARTS-C19$^\ddag$ & 19.15 & 1.010 & 1.34 \\
~ & P-DARTS$^\ddag$ \cite{chen2019progressive} & 15.67 & 3.485 & 4.58 \\
~ & PC-DARTS$^\ddag$ \cite{xu2019pc} & 16.66 & 3.687 & 4.29 \\
\cmidrule{2-5}
~ & E2RNAS-C38   & 19.30 & 3.459 & 4.90 \\
~ & E2RNAS-C36  &19.19   &3.120 & 4.00 \\
~ & E2RNAS-C29   & 19.80 & 2.075 & 3.78 \\
~ & E2RNAS-C20   & 22.97 & 1.041 & 3.44 \\
\midrule
\multirow{10}*{\rotatebox{90}{\textbf{SVHN}}} & DARTS$^\ddag$ \cite{lsy19} & 2.16 & 3.449 & 46.78 \\
~ & DARTS-C34$^\ddag$ & 2.18 & 2.998 & 41.32 \\
~ & DARTS-C28$^\ddag$ & 2.13 & 2.061 & 35.35 \\
~ & DARTS-C20$^\ddag$ & 2.16 & 1.083 & 40.38 \\
~ & P-DARTS$^\ddag$ \cite{chen2019progressive} & 2.12 & 3.433 & 49.11 \\
~ & PC-DARTS$^\ddag$ \cite{xu2019pc} & 2.20 & 3.635 & 54.81 \\
\cmidrule{2-5}
~ & E2RNAS-C39   & 2.21 & 3.421 & 44.15 \\
~ & E2RNAS-C36 & 2.14 & 2.935 & 53.82 \\
~ & E2RNAS-C30   & 2.13 & 2.075 & 52.38 \\
~ & E2RNAS-C21   & 2.21 & 1.062 & 54.96 \\
\bottomrule
\end{tabular}}
\vskip 0.05in
\caption{Comparison with state-of-the-art NAS methods on the CIFAR-100 and SVHN datasets. $\ddag$ indicates the use of the code released by original authors. $\uparrow$ indicates larger value is better, while $\downarrow$ indicates lower value is better. ``$\{\text{model-C}\#\text{channels}\}$" means the architecture searched by ``model" is evaluated with the initial number of channels as ``channels".}
\label{tab:result_cifar100_svhn}
\end{table}

\begin{table*}[!htbp]
\centering
% \resizebox{\linewidth}{!}{
\begin{tabular}{l|cccc|ccc}
\toprule
\multirow{2}*{\textbf{Method}} & \multirow{2}*{\texttt{adv}} & \multirow{2}*{\texttt{nop}} & \multirow{2}*{\texttt{MGDA}} & \texttt{L} & \textbf{Test Err.} & \textbf{Params}  & \textbf{PGD Acc.}\\
~ & ~ & ~ & ~ & \text{(MB)} & \textbf{(\%)}~$\downarrow$ & \textbf{(MB)}~$\downarrow$ & \textbf{(\%)}~$\uparrow$ \\
\midrule
E2RNAS & $\surd$ & $\surd$ & $\surd$ & $1$ &  4.19 & 2.102 & 9.61 \\
\midrule
\quad w/o~~\texttt{adv} & ~ & $\surd$ & $\surd$ & $1$ & 2.75 & 3.733 & 10.35 \\
\quad w/o~~\texttt{adv}~(C27) & ~ & $\surd$ & $\surd$ & $1$ & 2.84 & 2.148 & 8.91 \\
\quad w/o~~\texttt{adv}~($L=0$) & ~ & $\surd$ & ~ & $0$ & 7.95 & 1.370 & 4.00 \\
\quad w/o~~\texttt{nop} & $\surd$ & ~ & ~ & ~ & 8.29 & 1.370 & 5.21 \\
\quad w/o~~\texttt{MGDA} & $\surd$ & $\surd$ & ~ & $1$ & 5.48 & 2.105 & 8.11 \\
\quad w/o~~\texttt{L} & $\surd$ & $\surd$ & $\surd$ & ~ & 8.30 & 1.370 & 4.39 \\
% \midrule
% DARTS & ~ & ~ & ~ & ~ & 2.59 & 3.349 & 6.57 \\
\bottomrule
\end{tabular}
\vskip 0.05in
\caption{Ablation study on the CIFAR-10 dataset. \texttt{adv} means using adversarial training in the low-level problem of problem (\ref{eq:obj_fun}); \texttt{nop} indicates adding the resource constraint into the upper-level problem of problem (\ref{eq:obj_fun}); \texttt{MGDA} denotes using MGDA to make a trade-off between the accuracy and model size and if without \texttt{MGDA}, it means equal weights of the two objectives in problem (\ref{eq:obj_fun}) are used (\ie $\hat{\gamma}\equiv0.5$ in Eq. (\ref{eq:alpha_update})). \texttt{L} is the lower bound of the number of parameters. $\uparrow$ indicates larger value is better, while $\downarrow$ indicates lower value is better. ``C27" means the initial number of channels in the architecture evaluation is changed to 27, instead of the default number of 36.}
\label{tab:ablation_study}
\end{table*}

Compared to the original DARTS in \cite{lsy19}, ``E2RNAS-C36" significantly improves the robustness with lower model size and comparable search cost while the classification error increases slightly.  %Empirically, more parameter-free operations in network architecture often result in worse accuracy and robustness.
Some studies \cite{raghunathan2019adversarial, yang2020closer} show that the increased robustness is usually accompanied by decreased test accuracy. Therefore, the increased test error of E2RNAS is because of the improved robustness and the decreased parameter size, which indicates E2RNAS can make a better trade-off among these three goals than DARTS. %We will conduct further discussion in ablation study (Section \ref{sec:ablation}).

Besides, both P-DARTS \cite{chen2019progressive} and PC-DARTS \cite{xu2019pc} search for a deeper architecture with less search cost than ``E2RNAS-C36", so they slightly outperform in the test error rate with competitive PGD accuracy. We can apply the E2RNAS method to P-DARTS and PC-DARTS to make a trade-off among multiple objectives (\ie the accuracy, the robustness and the number of parameters) in future work.

To further compare the performance of E2RNAS and DARTS, we change the initial number of channels in the architecture evaluation for both methods to keep a roughly similar model size. According to the results shown in Table \ref{tab:result_cifar10}, we can see that E2RNAS remarkably improves the robustness with comparable classification accuracy. For example, compared ``E2RNAS-C46" with ``DARTS", the PGD accuracy increases about 1.6 times, while the test error increases by only around 0.9\%.

In summary, experiment results in Table \ref{tab:result_cifar10} show tat E2RNAS can search significantly robust architectures with a lower model size and comparable classification accuracy, compared with state-of-the-art NAS methods.
% For example, even the PGD accuracy of E2RNAS with only 0.449MB parameter size is higher than DARTS with 3.349MB model size (\ie ``E2RNAS-C16" \vs ``DARTS" in Table \ref{tab:result_cifar10}).

\vspace{-0.3cm}
\paragraph{Architecture Evaluation on CIFAR-100 and SVHN}
The comparison of E2RNAS with DARTS on the CIFAR-100 and SVHN datasets is presented in Table \ref{tab:result_cifar100_svhn}. The performance of E2RNAS on the CIFAR-100 dataset is similar to that on the CIFAR-10 dataset in that E2RNAS can search a robust architecture with a lower model size and a slightly decreased test accuracy. For example, compared to DARTS, ``E2RNAS-C36" reduces the number of parameters by 0.3MB and improves the PGD accuracy by nearly twice times, though the test error is slightly increased (about 2\%). In addition, E2RNAS shows excellent results on the SVHN dataset. It not only significantly improves the robustness but also achieves competitive test accuracy with a lower parameter size. For instance, compared to DARTS, ``E2RNAS-C36" reduces the model size by about 15\% and increases the PGD accuracy, while keeping competitive performance. Therefore, those quantitative experiments indicate that E2RNAS can search robust architectures with a lower model size and comparable performance.

% \subsection{Generalization Analysis of E2RNAS}

\begin{figure*} [!htb]
%\vskip -0.1in
  \centering
    \subfigure[$L$ \vs Test Error]{\includegraphics[width=0.33\textwidth]{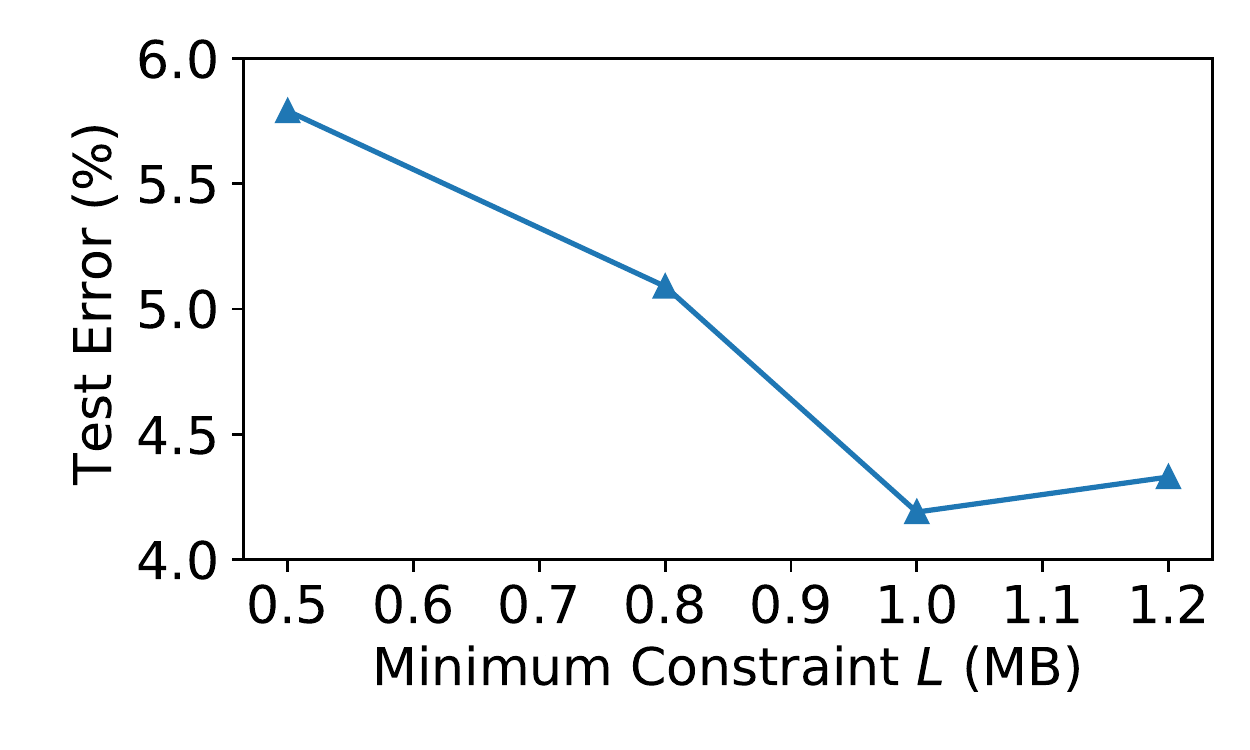}{\label{fig:L_err}}}
    \subfigure[$L$ \vs The Number of Parameters]{\includegraphics[width=0.33\textwidth]{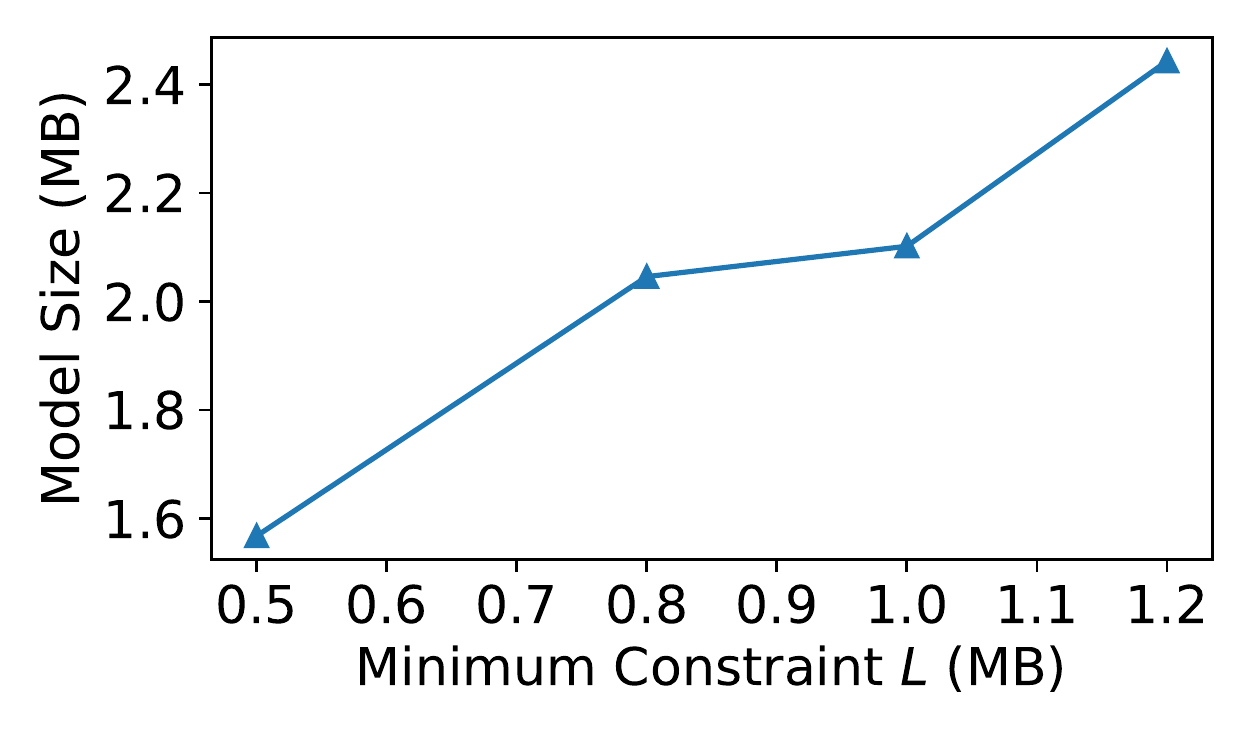}{\label{fig:L_nop}}}
    \subfigure[$L$ \vs PGD Accuracy]{\includegraphics[width=0.33\textwidth]{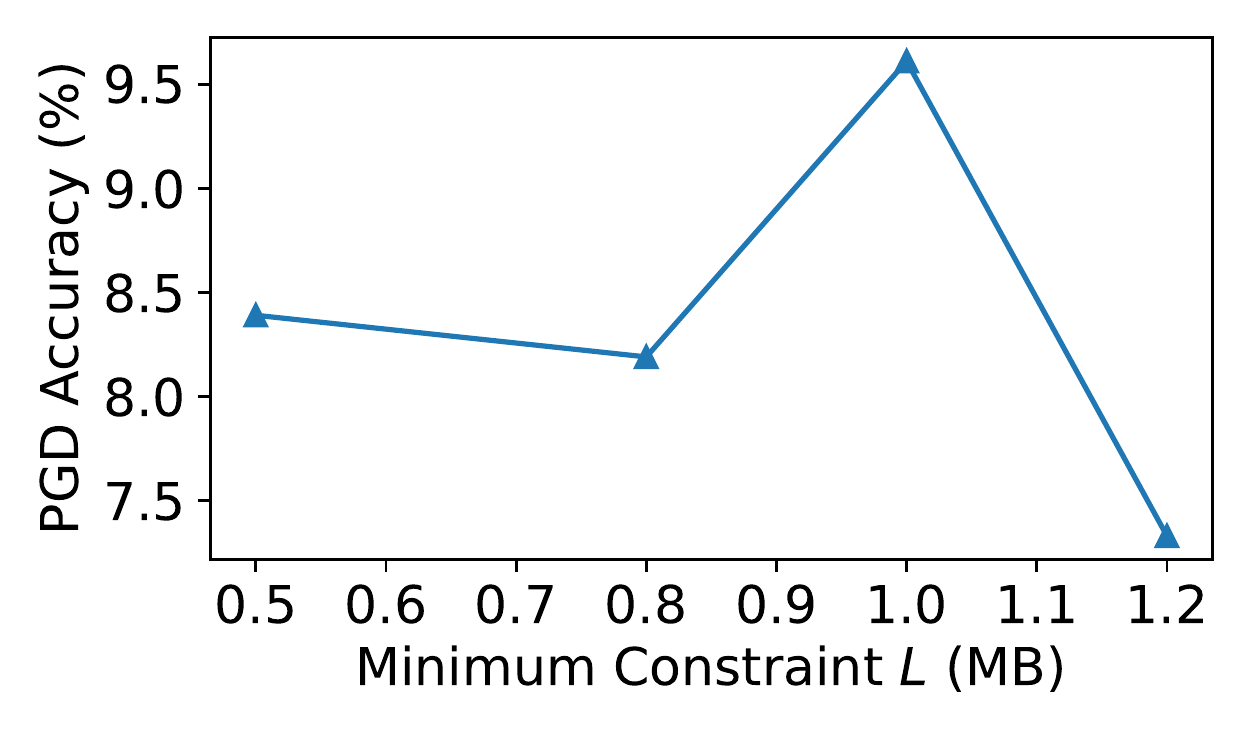}{\label{fig:L_pgd}}}
  \caption{Architecture evaluation of E2RNAS on the CIFAR-10 dataset using different minimum constraint $L$ in Eq. (\ref{eq:num_param_loss}).}
\label{fig:L_err_nop_pgd}
\end{figure*}

\subsection{Ablation Study} \label{sec:ablation}

In this section, we study how each design in E2RNAS influences its performance on different objectives. The corresponding results are presented in Table \ref{tab:ablation_study}. The adversarial training (abbreviated as \texttt{adv}) in the lower-level problem of problem (\ref{eq:obj_fun}) transforms training data to adversarial examples and hopes to learn a robust model when given an architecture. The resource constraint (abbreviated as \texttt{nop}) in the upper-level problem of problem (\ref{eq:obj_fun}) expects to constrain the parameter size of the searched architecture. The multiple-gradient descent algorithm (abbreviated as \texttt{MGDA}) is applied to solve the upper-level problem of problem (\ref{eq:obj_fun}), which is a multi-objective problem to minimize both the validation accuracy and the model size. If without \texttt{MGDA}, it means that we solve the upper-level problem by minimizing an equally weighted sum of two objectives (\ie $\hat{\gamma}\equiv0.5$ in Eq. (\ref{eq:alpha_update})). The lower bound (abbreviated as \texttt{L}) of the number of parameters expects to prevent the model to search over-simplified architectures.

\vspace{-0.3cm}
\paragraph{Impact of Adversarial Training}
The adversarial training, which trains a neural network on adversarial examples, is an effective method for improving the robustness of a neural network. Thus, we apply it in the lower-level problem of problem (\ref{eq:obj_fun}) and hope the searched architecture can defense adversarial attacks. Here we discuss two impacts of the adversarial training in details.

Firstly, using the adversarial training tends to reduce the number of parameters, which may leads to worse accuracy. We notice that the parameter size of the architecture searched by DARTS with adversarial training (\ie ``E2RNAS w/o \texttt{nop}" in Table \ref{tab:ablation_study}) is only 1.37MB, which means that the searched architecture contains many parameter-free operations. Therefore, it has a larger test error because of its simplistic architecture, although its PGD accuracy is larger than DARTS with a comparable model size (\ie ``DARTS-C20" in Table \ref{tab:result_cifar10}). Besides, compared with E2RNAS, the model size of ``E2RNAS w/o \texttt{adv}" increases by 1.631MB, which indicates that the adversarial training significantly decreases the number of parameters. However, it can be alleviated by constraining the parameter size with a lower bound $L$.

Secondly, using the adversarial training can help E2RNAS make a trade-off between the robustness and accuracy. We notice adversarial training can significantly influences the model size. Therefore, to make a fair comparison, we set the number of initial channels of ``E2RNAS w/o \texttt{adv}" (\ie ``E2RNAS w/o \texttt{adv} (C27)") in the architecture evaluation to keep its model size roughly similar to E2RNAS. The result in Table \ref{tab:ablation_study} shows ``E2RNAS" has a better robustness but lower accuracy than ``E2RNAS w/o \texttt{adv}~(C27)".

Therefore, using adversarial training can help E2RNAS to make a trade-off among multiple objectives and search a robust architecture with a lower model size.

% Compared with E2RNAS, the model size of ``E2RNAS w/o \texttt{adv}~($L=0.5$)" is comparable but its PGD accuracy is significantly lower. This implies that E2RNAS equipped with the adversarial training can search a more robust architecture to defense adversarial attacks.

% Thirdly, applying the adversarial training sometimes decreases the classification accuracy due to the increasing of the robustness. We notice that the test error of ``E2RNAS w/o \texttt{adv}~($L=0.5$)" is lower than that of ``E2RNAS", although they have similar parameter size.
% On the other hand,

\vspace{-0.3cm}
\paragraph{Effectiveness of MGDA}
MGDA is used to solve the upper-level problem of problem (\ref{eq:obj_fun}). We quantitatively compare the performance of E2RNAS with and without MGDA (\ie ``E2RNAS" \vs ``E2RNAS w/o \texttt{MGDA}" in Table \ref{tab:ablation_study}) and find that solving with MGDA achieves much better results on the test accuracy, parameter size, and PGD accuracy. So instead of using equal weights, using MGDA can find a good solution of weights and make a trade-off among multiple objectives.  %Therefore, this experiment illustrates the effectiveness of MGDA.

\vspace{-0.3cm}
\paragraph{Necessity of $L$}
We find that training E2RNAS without the minimum constraint $L$ (\ie ``E2RNAS w/o \texttt{L}" in Table \ref{tab:ablation_study}) searches an architecture with many parameter-free operations (\ie its parameter size is only 1.370MB). There are three reasons for this phenomenon. Firstly, the instability of DARTS sometimes makes it converge to extreme architectures (\eg full of skip-connects) \cite{zela2019understanding, chen2019progressive}. Secondly, as discussed above, using the adversarial training in the lower-level problem tends to reduce the number of parameters. Finally, only optimizing the number of parameters in the upper-level problem (\ie ``E2RNAS w/o \texttt{adv}~($L=0$)" in Table \ref{tab:ablation_study}) also results in searched architectures with many parameter-free operations. Therefore, it is necessary to constrain the number of parameters with a lower bound $L$ to prevent E2RNAS to search over-simplified architectures.

Figure \ref{fig:L_err_nop_pgd} shows the architecture evaluation results of E2RNAS on the CIFAR-10 dataset using different $L$ in Eq. (\ref{eq:num_param_loss}). Hence, we set this hyperparameter $L$ to 1 in our work because E2RNAS achieves the best performance (\ie lowest test error rate in Figure \ref{fig:L_err},  acceptable model size in Figure \ref{fig:L_nop}, highest PGD accuracy in Figure \ref{fig:L_pgd}) when $L=1$.

% \begin{table*}[!htbp]
% \centering
% \begin{tabular}{lccc}
% \toprule
% \multirow{2}*{\textbf{Architecture}} & \textbf{Test Err. (\%)}  & \textbf{Params (MB)}  & \textbf{PGD Acc. (\%)}\\
% ~ & \downarrow & \downarrow & \uparrow \\
% \midrule
% P-DARTS \cite{chen2019progressive}   &2.59 &3.434 & 8.35    \\
% E2RNAS on P-DARTS      &-    &-     &-   \\
% \midrule
% PC-DARTS \cite{xu2019pc}   &2.65 &3.635 & 9.53    \\
% E2RNAS on PC-DARTS      &-    &-     &-   \\
% \bottomrule
% \end{tabular}
% \vskip 0.1in
% \caption{Comparison with state-of-the-art architectures on SVHN. $\ddag$ denotes our implementation according to author's released code.}
% \label{tab:result_p_pc_darts}
% \end{table*}

\section{Conclusions}
In this paper, we propose the E2RNAS method that optimizes multiple objectives simultaneously to search an effective, efficient and robust architecture. The proposed objective function is formulated as a bi-level multi-objective problem and we design an algorithm to integrate the MGDA with the bi-level optimization. Experiments demonstrate that E2RNAS can find adversarial robust architecture with optimized model size and  comparable classification accuracy on various datasets. In our future study, we are interested in extending the proposed E2RNAS method to search for multiple Pareto-optimal architectures at one time.%There are many directions of future work that we would like to explore. One issue that we do not mention in this work was searching for multiple Pareto optimal architectures at one time. We would like to extend our method to address this issue in the future.

% \clearpage

\bibliographystyle{ieee_fullname}
\bibliography{E2RNAS}

\end{document}